\pdfoutput=1

\documentclass[11pt]{article}

\usepackage[]{EMNLP2022}

\usepackage{times}
\usepackage{latexsym}

\usepackage[T1]{fontenc}

\usepackage[utf8]{inputenc}

\usepackage{microtype}

\usepackage{inconsolata}

\usepackage{amsmath,amssymb}
\usepackage{parskip}
\usepackage{graphicx}
\usepackage{setspace} 
\usepackage{tabu}
\usepackage{enumitem}
\usepackage{booktabs}

\usepackage{soul}

\usepackage{etoolbox}
\AtBeginEnvironment{quote}{\par\singlespacing\small}

\usepackage{multirow}

\usepackage{hyperref}
\hypersetup{
    colorlinks=true,
    filecolor=magenta,      
    urlcolor=blue,
    pdftitle={Summarization for Program Synthesis},
    pdfpagemode=FullScreen,
}
\usepackage{listings}
\usepackage{color}

\definecolor{dkgreen}{rgb}{0,0.6,0}
\definecolor{gray}{rgb}{0.5,0.5,0.5}
\definecolor{mauve}{rgb}{0.58,0,0.82}

\title{Less is More: Summary of Long Instructions is Better for\\Program Synthesis}

\author{First Author \\
  Affiliation / Address line 1 \\
  Affiliation / Address line 2 \\
  Affiliation / Address line 3 \\
  \texttt{email@domain} \\\And
  Second Author \\
  Affiliation / Address line 1 \\
  Affiliation / Address line 2 \\
  Affiliation / Address line 3 \\
  \texttt{email@domain} \\}
  
\author{Kirby Kuznia\thanks{~~Equal Contribution} \quad Swaroop Mishra\footnotemark[1] \quad Mihir Parmar \quad Chitta Baral \vspace{5mm} \\ Arizona State University}
\begin{document}
\maketitle

\begin{abstract}

Despite the success of large pre-trained language models (LMs) such as Codex, they show below-par performance on the larger and more complicated programming related questions. We show that LMs benefit from the summarized version of complicated questions. Our findings show that superfluous information often present in problem description such as human characters, background stories, and names (which are included to help humans in understanding a task) does not help models in understanding a task. To this extent, we create a meta-dataset from the frequently used APPS dataset and the newly created CodeContests dataset for the program synthesis task. Our meta-dataset consists of human and synthesized summaries of the long and complicated programming questions. Experimental results on Codex show that our proposed approach outperforms baseline by $8.13\%$ on the APPS dataset and $11.88\%$ on the CodeContests dataset on average in terms of strict accuracy. Our analysis shows that summaries significantly improve performance for introductory (9.86\%) and interview (11.48\%) programming questions. However, it shows improvement by a small margin ($\sim 2\%$) for competitive programming questions, implying scope for future research in this direction.\footnote{Code and data is available at \url{https://github.com/kurbster/Prompt-Summarization}}

\end{abstract}

\section{Introduction}


Recently, large pre-trained LMs have been proven pivotal in programming-related tasks~\cite{wang2021codet5, chen2021codex, hendrycksapps2021, lu2021codexglue, papineni2002bleu}\footnote{Detailed related work is presented in Appendix \ref{app:related_work}}. Program synthesis aims to generate a code given the natural language description of a problem. Programming requirements in these problems vary in terms of complexity from a 3-5 line simple function to multiple functions that use advanced data structures. However, LMs such as Codex show below-par performance on the long and complicated programming questions. We observe that the natural language description of the program becomes long and complicated when there is superfluous information (see section ~\ref{sec:HumanSummaries}). The goal of adding this information to the description is to make it more understandable to humans. However, we find that this information confuses the model in understanding a task\footnote{See example in Appendix ~\ref{app:example_fake}}. We propose that removing the excess information and providing the model with the exact specifications of the problem can improve the performance of the LMs.




To remove excess information\footnote{Instructions for creating summaries given in Appendix \ref{app:instructions}}, we summarize the descriptions of the program in such a way that it does not lose important specifications. We use the APPS dataset \cite{hendrycksapps2021} and CodeContests dataset \cite{li2022competition} which are a collection of coding problems from different online sources and create a meta-dataset consisting of human and synthesized summaries.

We perform all experiments using the GPT-based Codex model \cite{chen2021codex} on the proposed meta-dataset and show that the summarized version of complicated questions improves strict accuracy by 8.13\% on the APPS dataset and 11.85\% on CodeContests. From our analysis, we can see significant improvement for introductory (9.86\%) and interview (11.48\%) related programming questions. However, it shows improvement by a small margin ($\sim 2\%$) for competitive programming questions. Considering that automatic evaluation of a program does not reward for partial correctness, we perform qualitative evaluation on our meta-dataset and find that original questions often confuse models in understanding the underlying problem, as models latch on to some spurious words in the text (e.g. the word `list' in question makes the model design a list even though the underlying problem is on graphs). We further analyze model performance on different types of summaries (i.e., basic, expert, and synthetic) and provide instruction-design principles that can help future research on prompting in program synthesis.



\section{Method}

\subsection{Dataset}

We use the APPS \cite{hendrycksapps2021} and CodeContests \cite{li2022competition} datasets to create summaries. We crowd-sourced the creation of human summaries. The result was 373 human summaries for APPS and 80 summaries for CodeContests along with and 8663 synthetic summaries using both datasets. Table \ref{table:AllProblemNums} shows the statistics of the generated summaries.


\begin{table}[h]
    \centering 
    \setlength\tabcolsep{4.0pt}
    \footnotesize
    
\resizebox{0.85\linewidth}{!}{
\begin{tabular}{ccc}
\toprule
\textbf{Data Source} & \textbf{Difficulty} & \textbf{\# of Problems} \\
 \midrule
 \multirow{4}{*}{Human}
 & Introductory & 145 \\
 & Interview & 123 \\
 & Competition & 105 \\
 & CodeContests & 80 \\
 \midrule
 & Total & 453 \\
 \midrule
 \multirow{4}{*}{Studio21}
 & Introductory & 1588 \\
 & Interview & 4551 \\
 & Competition & 1286 \\
 & CodeContests & 80 \\
 \midrule
 & Total & 7505 \\
 \midrule
 \multirow{3}{*}{GPT-3}
 & Introductory & 194 \\
 & Interview & 267 \\
 & Competition & 244 \\
 & CodeContests & 80 \\
 \midrule
 & Total & 785 \\ 
 \midrule
 \multirow{4}{*}{PEGASUS}
 & Introductory & 145 \\
 & Interview & 123 \\
 & Competition & 105 \\
 \midrule
 & Total & 373 \\
 \bottomrule
\end{tabular}
}
\caption{Statistics of the proposed meta-dataset.}
\label{table:AllProblemNums}
\end{table}

\subsubsection{Human Generated Summaries}
\label{sec:HumanSummaries}

For the APPS and CodeContests human-generated summaries, the crowd worker reads and understands the original questions, then creates summaries in two steps\footnote{Instructions for creating summaries are in Appendix \ref{app:instructions}}. First, we create a basic summary of the given problem and remove any information that is repeated and any hypothetical information without concrete instructions. For example, if the problem constructs a fake company or situation, 
we replace the fake situation with direct instructions. Full example is included in Appendix \ref{app:example_fake}. Second, we create an expert summary of the problem. To create this, we further summarize the first summary. This expert summary includes the absolute minimum information for an expert to understand the problem. We would not expect a novice to understand these prompts. An example of expert summaries is given in Appendix \ref{app:expert}.


\subsubsection{Synthetic Summaries}

We have generated synthetic summaries of program descriptions using jumbo (178B), large (7.5B) Studio21 model \cite{J1WhitePaper}, GPT-3 Davinci model (175B) \cite{brown2020language} and PEGASUS model \cite{zhang2019pegasus}. To generate a summary, we provide these models with a few examples in the in-context learning setup \cite{brown2020language} from the human-generated summaries. For the few-shot examples, we use expert-level summaries. 

\paragraph{Studio21} We use five examples with the large model, and three examples with the jumbo model\footnote{Examples are included in Appendix \ref{app:prompts}}. For both models, we use a temperature of $0.3$, and topP of $1$. For the format of our prompt, we use De-Jargonizer template\footnote{\url{https://studio.ai21.com/}} with a change to their header as shown in Appendix \ref{app:prompts}. We create a total of $7,505$ synthetic summaries using these models.


\paragraph{GPT-3} We use three examples for GPT-3 model. We empirically set temperature to $0.05$, topP to $1$, frequency penalty to $0.01$, presence penalty to $0.05$. To generate prompts, we followed their tl;dr template\footnote{\url{https://beta.openai.com/playground/p/default-tldr-summary?model=text-davinci-001}} as shown in Appendix \ref{app:prompts}. We create $785$ synthetic summaries using this model.

\paragraph{PEGASUS} We use the PEGASUS model \cite{zhang2019pegasus} to create program summaries for the same set of problems that were summarized by humans. We choose this model because it was trained specifically for abstractive summarization.





\begin{table*}[t!]
    \centering 
    \setlength\tabcolsep{4.0pt}
    \footnotesize
\resizebox{0.95\linewidth}{!}{

\begin{tabular}{cccc|ccc|ccc}
\toprule
\multirow{2}{*}{Difficulty} & \multicolumn{3}{c}{AP} & \multicolumn{3}{c}{EWPR} & \multicolumn{3}{c}{BWPR} \\ 
\cmidrule(lr){2-4}\cmidrule(lr){5-7}\cmidrule(lr){8-10}
~ & Baseline & Basic & Expert & Baseline & Basic & Expert & Baseline & Basic & Expert \\ 
 \midrule 
 Introductory & 42.96 & \textbf{50.00} & \textbf{50.00} &  44.20 & 51.45 & \textbf{51.82} & 43.23 & \textbf{50.35} & \textbf{50.35} \\ 
 Interview & 37.70 & 41.80 & \textbf{44.26} & 36.52 & 45.54 & \textbf{46.96} & 37.70 & 41.80 & \textbf{44.26} \\
 Competition & 4.76 & \textbf{5.71} & \textbf{5.71} & 4.00 & \textbf{6.00} & \textbf{6.00} & 4.76 & \textbf{5.71} & \textbf{5.71} \\
 \midrule
 Weighted Average & 30.47 & 34.83 & \textbf{35.64} & 30.31 & 36.65 & \textbf{37.22} & 30.43 & 34.78 & \textbf{35.59} \\
 \midrule
  CodeContests & 12.50 & 23.75 & \textbf{25.00} & 13.33 & 25.33 & \textbf{26.66} & 12.82 & 24.36 & \textbf{25.64} \\
 \bottomrule
\end{tabular}
}
\caption{Results of baseline and proposed model in terms of Strict Accuracy (SAcc). The first block is from the APPS dataset. The last block is from the CodeContests dataset. AP: All Problems, EWPR: Either Worst Problem Removal, BWPR: Both Worst Problem Removal (see explanation in section \ref{sec:exp-setup}). All results are in \%. Weighted Average is not shown for CodeContests because similar difficulties were not provided (see explanation in \ref{sec:human-results}).}
\label{tab:main-results}
\end{table*}



\begin{table}[t!]
\centering 
\setlength\tabcolsep{4.0pt}
\footnotesize
\resizebox{\linewidth}{!}{
\begin{tabular}{ccc|cc}
\toprule
\multirow{2}{*}{Difficulty} & \multicolumn{2}{c}{AP} & \multicolumn{2}{c}{EWPR} \\
\cmidrule(lr){2-3}\cmidrule(lr){4-5}
~ & Baseline & Proposed & Baseline & Proposed \\ 
 \midrule
 Introductory & 42.96 & \textbf{52.82} & 44.53 & \textbf{54.74} \\
 Interview & 37.70 & \textbf{49.18} & 38.66 & \textbf{50.42} \\
 Competition & 4.76 & \textbf{6.67} & 4.81 & \textbf{6.73} \\
 \midrule
 Weighted Average & 30.47 & \textbf{38.48} & 31.11 & \textbf{39.44} \\
 \bottomrule
\end{tabular}
}
\caption{Results when taking the best summary for each problem. The EWPR baseline is different from Table \ref{tab:main-results} because a different set of problems have been removed.}
\label{tab:take-best}
\end{table}

\subsection{Model}

We use OpenAI Codex to build baselines and the proposed approach.

\paragraph{Baseline} To create a baseline, we have used original program descriptions given in the datasets as prompts for the Codex model.

\paragraph{Proposed Approach} We have used summaries of original program descriptions given in the datasets as prompts for the Codex model.


\section{Experimental Setup}
\label{sec:exp-setup}
All the experiments are performed using the $davinci-codex$ \cite{chen2021codex} model provided through OpenAI\footnote{Implementation and parameters details in Appendix \ref{app:codex}}. At inference time, we use a modified version of the evaluation code\footnote{\url{https://github.com/hendrycks/apps/blob/main/eval/test_one_solution.py}} provided by \citet{hendrycksapps2021}. This evaluation code has four different outputs for each test case: (1) \textbf{-2}: the code has a syntax error and can not run, (2) \textbf{-1}: the code is syntactically correct but has a run time error, (3) \textbf{0}: the code runs without any errors but fails the test case, and (4) \textbf{1}: the code runs without any error and passes the test case. Similar to \citet{chen2021codex}, we implement a timeout for the code at inference time. If a test case takes more than $4$ seconds to run then we throw an exception and count that test case as a $-1$.


\paragraph{Experiments} To show effectiveness of the proposed approach, we have performed three different experiments using human generated summaries:
\begin{enumerate}[itemsep=-1ex]
    \item All problems from basic and expert summaries are used at inference time. We term this experiment All Problems (AP).
    \item We eliminate problems that perform worse\footnote{Definition of the worst problem is given in Appendix \ref{app:worst_probs}} 
    for either basic or expert summaries. We term this experiment Either Worst Problem Removal (EWPR).
    \item We eliminate problems that perform worse 
    for both basic and expert summaries. We term this experiment Both Worst Problem Removal (BWPR).
\end{enumerate}



\paragraph{Motivation behind EWPR and BWPR}
If a summary caused every test case to perform worse then it's likely the crowd worker produced a faulty summary. To mitigate the effect of outliers in the dataset, we use the EWPR method to remove such problems. Another hypothesis is that every problem benefits from some level of summarization (i.e., basic or expert). To measure this, we use the BWPR method. From Table \ref{tab:number-of-probs} results, we identify that only 1 problem had both summaries (basic and expert) preform worse.



\paragraph{Metric} In \cite{ProgSynthesisLargeLangModels}, they show that the BLEU metric \cite{papineni2002bleu} does not correlate well with synthesis performance. Thus, we use Strict Accuracy (SAcc) as our evaluation metric for all experiments (see Appendix \ref{app:accuracy}).

\section{Results and Analysis}
\begin{table*}[ht]
\centering 
\setlength\tabcolsep{4.0pt}
\footnotesize
\resizebox{0.75\linewidth}{!}{

\begin{tabular}{cccc|cc}
\toprule
\multirow{2}{*}{Model} & \multirow{2}{*}{Difficulty} & \multicolumn{2}{c}{AP} & \multicolumn{2}{c}{EWPR} \\
\cmidrule(lr){3-4}\cmidrule(lr){5-6}
~& ~ & Baseline & Proposed & Baseline & Proposed \\ 
 \midrule
 \multirow{2}{*}{GPT-3}
 & Introductory & \textbf{41.75} & 38.66 & 41.11 & \textbf{41.67} \\
 & Interview & \textbf{20.30} & 18.80 & 18.18 & \textbf{20.66} \\
 & Competition & 2.87 & \textbf{3.28} & 2.73 & \textbf{3.64} \\
 \midrule
 & Weighted Average & \textbf{20.17} & 18.89 & 19.14 & \textbf{20.55} \\
 \midrule
 \multirow{3}{*}{Studio21}
 & Introductory & \textbf{39.53} & 31.63 & \textbf{39.04} & 36.36 \\
 & Interview & \textbf{12.28} & 11.00 & 10.57 & \textbf{12.37} \\
 & Competition & \textbf{1.67} & 1.21 & \textbf{1.38} & \textbf{1.38}\\
 \midrule
 & Weighted Average & \textbf{11.53} & 9.66 & 10.61 & \textbf{10.98} \\
 \midrule
 \multirow{3}{*}{PEGASUS}
 & Introductory & \textbf{42.96} & 34.48 & \textbf{44.26} & 40.98 \\
 & Interview & \textbf{37.70} & 10.57 & 14.29 & \textbf{21.56}   \\
 & Competition & \textbf{4.76} & 0.00 & \textbf{2.76} & 0.00 \\
 \midrule
 & Weighted Average & \textbf{30.47} & 16.88 & \textbf{25.50} & 24.73 \\
 \bottomrule
\end{tabular}
}
\caption{Results of baseline and proposed approach (All results are in \%). Summaries generated by GPT-3, Studio21, and PEGASUS used for inference from APPS.}
\label{tab:synthetic-results}
\end{table*}

\begin{table}[ht]
\centering 
\setlength\tabcolsep{4.0pt}
\footnotesize
\resizebox{\linewidth}{!}{

\begin{tabular}{ccc|cc}
\toprule
\multirow{2}{*}{Model} & \multicolumn{2}{c}{AP} & \multicolumn{2}{c}{EWPR} \\
\cmidrule(lr){2-3}\cmidrule(lr){4-5}
~ & Baseline & Proposed & Baseline & Proposed \\ 
 \midrule
 \multirow{1}{*}{GPT-3}
 & \textbf{12.50} & 10.0 & \textbf{18.75} & \textbf{18.75} \\
 \midrule
 \multirow{1}{*}{Studio21}
 & \textbf{12.50} & 8.75 & \textbf{22.5} & 20.0 \\
 \bottomrule
\end{tabular}
}
\caption{Results of baseline and proposed approach (All results are in \%). 80 summaries generated by GPT-3 and Studio21 used for inference from CodeContests.}
\label{tab:code-contests-synthetic-results}
\end{table}

\subsection{Human Generated Summaries}
\label{sec:human-results}

From Table \ref{tab:main-results}, we can observe that both the summary-based models show on average superior performance compared to baseline. In particular, when calculating results for every problem, basic and expert summary-based models outperform baseline by 4.34\% and 5.15\% on average for APPS dataset, respectively. Further analysis shows that the expert summary-based model shows improved performance by $\sim1\%$ compared to the basic summary-based model.

On the CodeContests dataset \cite{li2022competition}, we show an average improvement of $11.88\%$ in terms of SAcc. For this dataset, we did not separate the problems by difficulty. This is because the problems come from different sources and have different scales of difficulty. Thus, we did not report the SAcc when weighted by difficulty in Table \ref{tab:main-results}. 


Our analysis shows that many problems where the basic summary would fail, however, the expert summary would succeed and vice-versa. Thus, we choose the best summary for each problem after evaluating both summaries and then calculate the results for the best summaries. Table \ref{tab:take-best} shows results when taking the best summary for each problem for APPS dataset. We observe a 9.86\%, 11.48\%, and 1.91\% increase on SAcc for introductory, interview, and competition level problems, respectively.


\subsection{Synthetic Summaries}


Table \ref{tab:synthetic-results} and \ref{tab:code-contests-synthetic-results} show the results for baseline, synthetic summaries generated by GPT-3, Studio21 and PEGASUS in terms of SAcc for two experiments. For the AP experiment, we can observe that the performance of the baseline outperforms synthetic summary-based models. However, the proposed model shows an average similar performance compared to the baseline for the EWPR experiment. Moreover, Appendix \ref{app:abb_syn_results} shows the results for top 500 and top 1000 summaries from GPT-3 and Studio21, respectively.


\subsection{Analysis}

\paragraph{Why does eliminating the worst problems help?} From Tables \ref{tab:main-results}, we can observe that EWPR and BWPR have improved performance compared to AP for both human and synthetically generated summaries. By analyzing the summarized worst problems, we notice a difference in the summarization style which shows that these summaries are outliers and do not match the distribution of the other summaries. This can cause a problem in synthesizing a good program since the model loses important information. Hence, we believe that eliminating the worst problems improves model performance.

\paragraph{Is there any possible bias in the meta-dataset?} Recent studies shows that bias propagates in human-annotated datasets \cite{geva-etal-2019-modeling, parmar2022don}. Given that our summaries are also human-generated, there will be some bias in the dataset. 
Some details that are critical to one person can be trivial to others. In the context of generating expert summaries, assumptions about expert knowledge can vary. 
This bias causes drift in the dataset and hinders the model's performance. Similar to \citet{mishra2021cross}, we can provide a template for what is expected from the summary generator to reduce bias.

\paragraph{Why is competition accuracy low?}
We believe that these problems require multi-hop reasoning, even after summarization, which is still a challenge for language models.


\begin{figure}
    \centering
    \includegraphics[width=\linewidth]{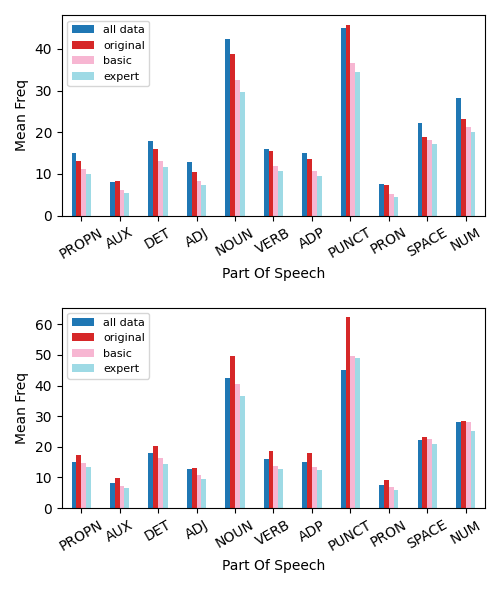}
    \caption{(Top plot) Mean frequency of POS for problems where programs where the generated by both the original and summarized prompt pass all test cases, and (Bottom plot) mean frequency of POS for problems where the summary passes all test cases and the original did not. The blue bar represents the mean of the entire dataset. analyzed only the top 11 most occurring POS. The plot shows that higher number of nouns degrade model performance.}
    \label{fig:pos-error-analysis}
\end{figure}

\paragraph{Impact of POS on Accuracy}
In the top plot of Figure \ref{fig:pos-error-analysis}, we observe that frequency of $nouns$ and $propernouns$ for problems that passed all test cases is lower than the entire dataset. In the bottom plot, we observe that the frequency for $nouns$ and $propernouns$ is higher for the original question (which had < 100\% accuracy on the test cases) and lower for the summary (which had 100\% accuracy on the test cases). Thus, we can see that number of nouns degrades performance. We also see in the bottom chart that overuse of punctuation can be detrimental to performance.
From the results in Figure \ref{fig:pos-error-analysis} we see results of nouns affecting performance along with excessive punctuation. 
Additional detailed analysis is presented in Appendix \ref{app:analysis}.

\section{Conclusion}

This paper introduces a summarization-based approach for efficient program synthesis. Experimental results show that the proposed approach improves the performance of the Codex model by on average $\sim8\%$ across various levels of programming questions provided by the APPS and $\sim11\%$ on the CodeContests. Further, this paper proposes a meta-dataset consisting of $\sim450$ human-generated basic and expert-level summaries as well as $\sim8k$ synthetically generated summaries by GPT-3 and Studio21; this can be helpful for future research on writing better instructions for the program synthesis. We show that program synthesis models benefit from concise prompts, hence, we believe that less number of high-quality instances are better than more low-quality data instances.

\paragraph{Future Extensions} The decomposition of prompts has been shown to improve accuracy \cite{mishra-etal-2022-reframing, patel2022question}; splitting up the summarization task the resulting summary can potentially result in higher accuracy for the Codex model in future. Additionally, the PEGASUS model could be used in conjunction with other models to perform the detailed algorithm outlined in Appendix \ref{app:instructions}.


\section*{Limitations}
Our summary-based approach shows improved performance on program synthesis models, however, it shows competitive performance on synthetic summaries. We believe that the generation of high-quality summaries can improve performance, hence, designing efficient prompts to improve synthetic summaries can be the scope of further research. Furthermore, human-generated summaries show competitive performance on competition-level problems. These problems require reasoning with multiple logical leaps and knowledge of advanced algorithms and data structures. Hence, exploring new techniques for summarization can be a future research direction. In addition, this work only analyzes the codex model, hence, exploring the effect of summarization on other program synthesis models can be interesting.



\bibliography{anthology,custom}

\begin{thebibliography}{28}
\expandafter\ifx\csname natexlab\endcsname\relax\def\natexlab#1{#1}\fi

\bibitem[{Austin et~al.(2021{\natexlab{a}})Austin, Odena, Nye, Bosma,
  Michalewski, Dohan, Jiang, Cai, Terry, Le
  et~al.}]{ProgSynthesisLargeLangModels}
Jacob Austin, Augustus Odena, Maxwell Nye, Maarten Bosma, Henryk Michalewski,
  David Dohan, Ellen Jiang, Carrie Cai, Michael Terry, Quoc Le, et~al.
  2021{\natexlab{a}}.
\newblock Program synthesis with large language models.
\newblock \emph{arXiv preprint arXiv:2108.07732}.

\bibitem[{Austin et~al.(2021{\natexlab{b}})Austin, Odena, Nye, Bosma,
  Michalewski, Dohan, Jiang, Cai, Terry, Le et~al.}]{austin2021program}
Jacob Austin, Augustus Odena, Maxwell Nye, Maarten Bosma, Henryk Michalewski,
  David Dohan, Ellen Jiang, Carrie Cai, Michael Terry, Quoc Le, et~al.
  2021{\natexlab{b}}.
\newblock Program synthesis with large language models.
\newblock \emph{arXiv preprint arXiv:2108.07732}.

\bibitem[{Balog et~al.(2016)Balog, Gaunt, Brockschmidt, Nowozin, and
  Tarlow}]{balog2016deepcoder}
Matej Balog, Alexander~L Gaunt, Marc Brockschmidt, Sebastian Nowozin, and
  Daniel Tarlow. 2016.
\newblock Deepcoder: Learning to write programs.
\newblock \emph{arXiv preprint arXiv:1611.01989}.

\bibitem[{Black et~al.(2021)Black, Gao, Wang, Leahy, and Biderman}]{GPTNeo}
Sid Black, Leo Gao, Phil Wang, Connor Leahy, and Stella Biderman. 2021.
\newblock Gpt-neo: Large scale autoregressive language modeling with
  mesh-tensorflow.
\newblock \emph{If you use this software, please cite it using these metadata},
  58.

\bibitem[{Brown et~al.(2020)Brown, Mann, Ryder, Subbiah, Kaplan, Dhariwal,
  Neelakantan, Shyam, Sastry, Askell et~al.}]{brown2020language}
Tom Brown, Benjamin Mann, Nick Ryder, Melanie Subbiah, Jared~D Kaplan, Prafulla
  Dhariwal, Arvind Neelakantan, Pranav Shyam, Girish Sastry, Amanda Askell,
  et~al. 2020.
\newblock Language models are few-shot learners.
\newblock \emph{Advances in neural information processing systems},
  33:1877--1901.

\bibitem[{Chen et~al.(2021)Chen, Tworek, Jun, Yuan, Pinto, Kaplan, Edwards,
  Burda, Joseph, Brockman et~al.}]{chen2021codex}
Mark Chen, Jerry Tworek, Heewoo Jun, Qiming Yuan, Henrique Ponde de~Oliveira
  Pinto, Jared Kaplan, Harri Edwards, Yuri Burda, Nicholas Joseph, Greg
  Brockman, et~al. 2021.
\newblock Evaluating large language models trained on code.
\newblock \emph{arXiv preprint arXiv:2107.03374}.

\bibitem[{Devlin et~al.(2017)Devlin, Uesato, Bhupatiraju, Singh, Mohamed, and
  Kohli}]{devlin2017robustfill}
Jacob Devlin, Jonathan Uesato, Surya Bhupatiraju, Rishabh Singh, Abdel-rahman
  Mohamed, and Pushmeet Kohli. 2017.
\newblock Robustfill: Neural program learning under noisy i/o.
\newblock In \emph{International conference on machine learning}, pages
  990--998. PMLR.

\bibitem[{Ge and Mooney(2005)}]{ge2005statistical}
Ruifang Ge and Raymond Mooney. 2005.
\newblock A statistical semantic parser that integrates syntax and semantics.
\newblock In \emph{Proceedings of the Ninth Conference on Computational Natural
  Language Learning (CoNLL-2005)}, pages 9--16.

\bibitem[{Geva et~al.(2019)Geva, Goldberg, and
  Berant}]{geva-etal-2019-modeling}
Mor Geva, Yoav Goldberg, and Jonathan Berant. 2019.
\newblock \href {https://doi.org/10.18653/v1/D19-1107} {Are we modeling the
  task or the annotator? an investigation of annotator bias in natural language
  understanding datasets}.
\newblock In \emph{Proceedings of the 2019 Conference on Empirical Methods in
  Natural Language Processing and the 9th International Joint Conference on
  Natural Language Processing (EMNLP-IJCNLP)}, pages 1161--1166, Hong Kong,
  China. Association for Computational Linguistics.

\bibitem[{Gulwani et~al.(2017)Gulwani, Polozov, Singh
  et~al.}]{gulwani2017program}
Sumit Gulwani, Oleksandr Polozov, Rishabh Singh, et~al. 2017.
\newblock Program synthesis.
\newblock \emph{Foundations and Trends{\textregistered} in Programming
  Languages}, 4(1-2):1--119.

\bibitem[{Hendrycks et~al.(2021)Hendrycks, Basart, Kadavath, Mazeika, Arora,
  Guo, Burns, Puranik, He, Song et~al.}]{hendrycksapps2021}
Dan Hendrycks, Steven Basart, Saurav Kadavath, Mantas Mazeika, Akul Arora,
  Ethan Guo, Collin Burns, Samir Puranik, Horace He, Dawn Song, et~al. 2021.
\newblock Measuring coding challenge competence with apps.
\newblock \emph{arXiv preprint arXiv:2105.09938}.

\bibitem[{Li et~al.(2022)Li, Choi, Chung, Kushman, Schrittwieser, Leblond,
  Eccles, Keeling, Gimeno, Lago et~al.}]{li2022competition}
Yujia Li, David Choi, Junyoung Chung, Nate Kushman, Julian Schrittwieser,
  R{\'e}mi Leblond, Tom Eccles, James Keeling, Felix Gimeno, Agustin~Dal Lago,
  et~al. 2022.
\newblock Competition-level code generation with alphacode.
\newblock \emph{arXiv preprint arXiv:2203.07814}.

\bibitem[{Lieber et~al.(2021)Lieber, Sharir, Lenz, and Shoham}]{J1WhitePaper}
Opher Lieber, Or~Sharir, Barak Lenz, and Yoav Shoham. 2021.
\newblock Jurassic-1: Technical details and evaluation.
\newblock \emph{White Paper. AI21 Labs}.

\bibitem[{Lu et~al.(2021)Lu, Guo, Ren, Huang, Svyatkovskiy, Blanco, Clement,
  Drain, Jiang, Tang et~al.}]{lu2021codexglue}
Shuai Lu, Daya Guo, Shuo Ren, Junjie Huang, Alexey Svyatkovskiy, Ambrosio
  Blanco, Colin Clement, Dawn Drain, Daxin Jiang, Duyu Tang, et~al. 2021.
\newblock Codexglue: A machine learning benchmark dataset for code
  understanding and generation.
\newblock \emph{arXiv preprint arXiv:2102.04664}.

\bibitem[{Luo et~al.(2022)Luo, Saxena, Mishra, Parmar, and
  Baral}]{luo2022biotabqa}
Man Luo, Sharad Saxena, Swaroop Mishra, Mihir Parmar, and Chitta Baral. 2022.
\newblock Biotabqa: Instruction learning for biomedical table question
  answering.
\newblock \emph{arXiv preprint arXiv:2207.02419}.

\bibitem[{Mishra et~al.(2022)Mishra, Khashabi, Baral, Choi, and
  Hajishirzi}]{mishra-etal-2022-reframing}
Swaroop Mishra, Daniel Khashabi, Chitta Baral, Yejin Choi, and Hannaneh
  Hajishirzi. 2022.
\newblock \href {https://doi.org/10.18653/v1/2022.findings-acl.50} {Reframing
  instructional prompts to {GPT}k{'}s language}.
\newblock In \emph{Findings of the Association for Computational Linguistics:
  ACL 2022}, pages 589--612, Dublin, Ireland. Association for Computational
  Linguistics.

\bibitem[{Mishra et~al.(2021)Mishra, Khashabi, Baral, and
  Hajishirzi}]{mishra2021cross}
Swaroop Mishra, Daniel Khashabi, Chitta Baral, and Hannaneh Hajishirzi. 2021.
\newblock Cross-task generalization via natural language crowdsourcing
  instructions.
\newblock \emph{arXiv preprint arXiv:2104.08773}.

\bibitem[{Nye et~al.(2021)Nye, Andreassen, Gur-Ari, Michalewski, Austin,
  Bieber, Dohan, Lewkowycz, Bosma, Luan et~al.}]{nye2021show}
Maxwell Nye, Anders~Johan Andreassen, Guy Gur-Ari, Henryk Michalewski, Jacob
  Austin, David Bieber, David Dohan, Aitor Lewkowycz, Maarten Bosma, David
  Luan, et~al. 2021.
\newblock Show your work: Scratchpads for intermediate computation with
  language models.
\newblock \emph{arXiv preprint arXiv:2112.00114}.

\bibitem[{Papineni et~al.(2002)Papineni, Roukos, Ward, and
  Zhu}]{papineni2002bleu}
Kishore Papineni, Salim Roukos, Todd Ward, and Wei-Jing Zhu. 2002.
\newblock Bleu: a method for automatic evaluation of machine translation.
\newblock In \emph{Proceedings of the 40th annual meeting of the Association
  for Computational Linguistics}, pages 311--318.

\bibitem[{Parmar et~al.(2022{\natexlab{a}})Parmar, Mishra, Geva, and
  Baral}]{parmar2022don}
Mihir Parmar, Swaroop Mishra, Mor Geva, and Chitta Baral. 2022{\natexlab{a}}.
\newblock Don't blame the annotator: Bias already starts in the annotation
  instructions.
\newblock \emph{arXiv preprint arXiv:2205.00415}.

\bibitem[{Parmar et~al.(2022{\natexlab{b}})Parmar, Mishra, Purohit, Luo,
  Mohammad, and Baral}]{parmar2022boxbart}
Mihir Parmar, Swaroop Mishra, Mirali Purohit, Man Luo, Murad Mohammad, and
  Chitta Baral. 2022{\natexlab{b}}.
\newblock \href {https://doi.org/10.18653/v1/2022.findings-naacl.10}
  {In-{B}o{XBART}: Get instructions into biomedical multi-task learning}.
\newblock In \emph{Findings of the Association for Computational Linguistics:
  NAACL 2022}, pages 112--128, Seattle, United States. Association for
  Computational Linguistics.

\bibitem[{Patel et~al.(2022)Patel, Mishra, Parmar, and
  Baral}]{patel2022question}
Pruthvi Patel, Swaroop Mishra, Mihir Parmar, and Chitta Baral. 2022.
\newblock Is a question decomposition unit all we need?
\newblock \emph{EMNLP 2022, Abu Dhabi}.

\bibitem[{Puri et~al.(2022)Puri, Mishra, Parmar, and Baral}]{puri2022many}
Ravsehaj~Singh Puri, Swaroop Mishra, Mihir Parmar, and Chitta Baral. 2022.
\newblock How many data samples is an additional instruction worth?
\newblock \emph{arXiv preprint arXiv:2203.09161}.

\bibitem[{Sanh et~al.(2021)Sanh, Webson, Raffel, Bach, Sutawika, Alyafeai,
  Chaffin, Stiegler, Scao, Raja et~al.}]{sanh2021multitask}
Victor Sanh, Albert Webson, Colin Raffel, Stephen~H Bach, Lintang Sutawika,
  Zaid Alyafeai, Antoine Chaffin, Arnaud Stiegler, Teven~Le Scao, Arun Raja,
  et~al. 2021.
\newblock Multitask prompted training enables zero-shot task generalization.
\newblock \emph{arXiv preprint arXiv:2110.08207}.

\bibitem[{Wang et~al.(2021)Wang, Wang, Joty, and Hoi}]{wang2021codet5}
Yue Wang, Weishi Wang, Shafiq Joty, and Steven~CH Hoi. 2021.
\newblock Codet5: Identifier-aware unified pre-trained encoder-decoder models
  for code understanding and generation.
\newblock \emph{arXiv preprint arXiv:2109.00859}.

\bibitem[{Wei et~al.(2021)Wei, Bosma, Zhao, Guu, Yu, Lester, Du, Dai, and
  Le}]{wei2021finetuned}
Jason Wei, Maarten Bosma, Vincent~Y Zhao, Kelvin Guu, Adams~Wei Yu, Brian
  Lester, Nan Du, Andrew~M Dai, and Quoc~V Le. 2021.
\newblock Finetuned language models are zero-shot learners.
\newblock \emph{arXiv preprint arXiv:2109.01652}.

\bibitem[{Wei et~al.(2022)Wei, Wang, Schuurmans, Bosma, Chi, Le, and
  Zhou}]{wei2022chain}
Jason Wei, Xuezhi Wang, Dale Schuurmans, Maarten Bosma, Ed~Chi, Quoc Le, and
  Denny Zhou. 2022.
\newblock Chain of thought prompting elicits reasoning in large language
  models.
\newblock \emph{arXiv preprint arXiv:2201.11903}.

\bibitem[{Zhang et~al.(2019)Zhang, Zhao, Saleh, and Liu}]{zhang2019pegasus}
Jingqing Zhang, Yao Zhao, Mohammad Saleh, and Peter~J. Liu. 2019.
\newblock \href {http://arxiv.org/abs/1912.08777} {Pegasus: Pre-training with
  extracted gap-sentences for abstractive summarization}.

\end{thebibliography}
\bibliographystyle{acl_natbib}

\clearpage

\appendix

\section{Related Work}
\label{app:related_work}

In the past, there are several methods including semantic parsing \cite{ge2005statistical}, deductive approaches, enumerative and stochastic search, and constraint solving which have gained attention for program synthesis \cite{gulwani2017program}. With the advent of machine/deep learning, \citet{balog2016deepcoder} introduced a neural network based model for solving programming competition-style problems. \citet{devlin2017robustfill} used sequence-to-sequence approach to do program synthesis. Furthermore, \citet{hendrycksapps2021} introduced the APPS dataset for testing the accuracy of large LMs on program synthesis. 
\citet{hendrycksapps2021} leveraged the \textit{GPT-Neo} model \cite{GPTNeo} which they fine-tune for this task using APPS dataset. CodeT5 model \cite{wang2021codet5} utilizes many different training objectives. Recently, \citet{austin2021program} explore limitations of large language models and propose two new benchmarks, MBPP and MathQA-Python. The Codex model \cite{chen2021codex} is an advanced code generation model that powers GitHub's \href{https://copilot.github.com/}{Copilot}. The state of the art model for program synthesis was introduced by Deepmind called AlphaCode \cite{li2022competition}. They released their dataset CodeContests, which was used to fine-tune and test their model, and was used in this paper. Our approach suggesting smaller instructions compliments other approaches in improving model performance in instruction paradigm~\cite{mishra2021cross, wei2022chain, parmar2022boxbart, nye2021show,  puri2022many, luo2022biotabqa, wei2021finetuned, sanh2021multitask}

\section{Additional Analysis}
\label{app:analysis}
\begin{figure}
    \centering
    \includegraphics[width=\linewidth]{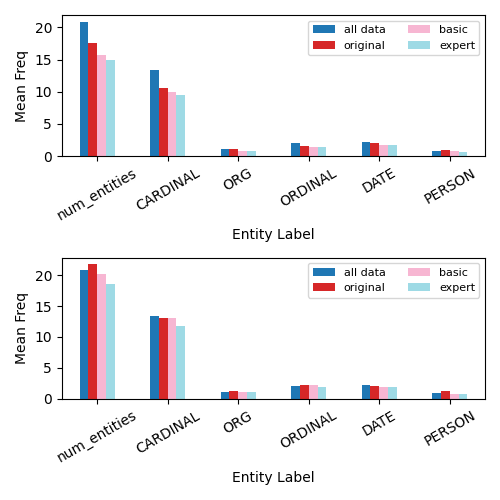}
    \caption{(Top plot) Mean frequency of the entity labels for problems where program generated by the original and summarized prompt pass all test cases, and (Bottom plot) mean frequency of entity labels for problems where the summary passes all test cases and the original did not. We analyzed only the top 5 most occurring entities among all entities we found.}
    \label{fig:ent-error-analysis}
\end{figure}

\paragraph{Difficulty of CodeContests}
The accuracies for CodeContests is notably lower than the APPS dataset since this dataset is more challenging, e.g. the number and complexity of programming operations is relatively higher than APPS. From the baseline results in Table \ref{tab:main-results}, we can observe that problems in CodeContests are harder than interview but easier than competition. 
\paragraph{Impact of Entities on Accuracy}
In Figure \ref{fig:ent-error-analysis}, we can observe that the total number of entities $num\_entities$ is higher for problems that performed worse. Here, we can see that the original problems (which failed test cases) had a higher mean than the dataset and the summaries (which passed all test cases) had a lower number of entities. 

\section{Example of removing fake information}
\label{app:example_fake}

To see the code produced by the model for this example, refer to Appendix \ref{app:generated-code}. There are more examples of superfluous information confusing the model in Appendix \ref{app:extra-information} and of made up information confusing the model in Appendix \ref{app:fake-information}.

\subsection{Original Prompt}

\begin{quote}
    Codefortia is a small island country located somewhere in the West Pacific. It consists of $n$ settlements connected by $m$ bidirectional gravel roads. Curiously enough, the beliefs of the inhabitants require the time needed to pass each road to be equal either to $a$ or $b$ seconds. It's guaranteed that one can go between any pair of settlements by following a sequence of roads.

Codefortia was recently struck by the financial crisis. Therefore, the king decided to abandon some of the roads so that:

  it will be possible to travel between each pair of cities using the remaining roads only,  the sum of times required to pass each remaining road will be minimum possible (in other words, remaining roads must form minimum spanning tree, using the time to pass the road as its weight),  among all the plans minimizing the sum of times above, the time required to travel between the king's residence (in settlement $1$) and the parliament house (in settlement $p$) using the remaining roads only will be minimum possible. 

The king, however, forgot where the parliament house was. For each settlement $p = 1, 2, \dots, n$, can you tell what is the minimum time required to travel between the king's residence and the parliament house (located in settlement $p$) after some roads are abandoned?

-----Input-----

The first line of the input contains four integers $n$, $m$, $a$ and $b$ ($2 \leq n \leq 70$, $n - 1 \leq m \leq 200$, $1 \leq a < b \leq 10^7$) — the number of settlements and gravel roads in Codefortia, and two possible travel times. Each of the following lines contains three integers $u, v, c$ ($1 \leq u, v \leq n$, $u \neq v$, $c \in \{a, b\}$) denoting a single gravel road between the settlements $u$ and $v$, which requires $c$ minutes to travel.

You can assume that the road network is connected and has no loops or multiedges.

-----Output-----

Output a single line containing $n$ integers. The $p$-th of them should denote the minimum possible time required to travel from $1$ to $p$ after the selected roads are abandoned. Note that for each $p$ you can abandon a different set of roads.

-----Examples-----

Input\\
5 5 20 25\\
1 2 25\\
2 3 25\\
3 4 20\\
4 5 20\\
5 1 20\\

Output\\
0 25 60 40 20

Input\\
6 7 13 22\\
1 2 13\\
2 3 13\\
1 4 22\\
3 4 13\\
4 5 13\\
5 6 13\\
6 1 13\\

Output\\
0 13 26 39 26 13\\

-----Note-----

The minimum possible sum of times required to pass each road in the first example is $85$ — exactly one of the roads with passing time $25$ must be abandoned. Note that after one of these roads is abandoned, it's now impossible to travel between settlements $1$ and $3$ in time $50$.
\end{quote}

We can see the author of the problem is trying to describe a fully-connected graph with $n$ nodes and $m$ edges each with a weight $a$ or $b$. Thus, this paragraph can be summarized as:

\subsection{Basic Summary}

\begin{quote}
You are given a graph of $n$ nodes and $m$ bidirectional edges.
The cost for each edge is either $a$ or $b$.
The graph is fully-connected, so you can travel between any pair of nodes using the edges. 

For each node $p = 1, 2, \dots, n$, you need to remove some edges so that:
It will be possible to travel between each pair of nodes using the remaining edges only, and
the sum of times required to pass each remaining road will be the minimum possible.
You should output the minimum time required to travel between node $1$ and node $p$.

-----Input-----

The first line of the input contains four integers $n$, $m$, $a$ and $b$ ($2 \leq n \leq 70$, $n - 1 \leq m \leq 200$, $1 \leq a < b \leq 10^7$) — the number of nodes and edges in the graph, and two possible travel times. Each of the following lines contains three integers $u, v, c$ ($1 \leq u, v \leq n$, $u \neq v$, $c \in \{a, b\}$) denoting an edge between the nodes $u$ and $v$, which has cost $c$.

You can assume that the graph is connected and has no loops or multiedges.

-----Output-----

Output a single line containing $n$ integers. The $p$-th of them should denote the minimum possible post required to travel from $1$ to $p$ after the selected edges are abandoned. Note that for each $p$ you can abandon a different set of edges.

-----Examples-----

Input\\
5 5 20 25\\
1 2 25\\
2 3 25\\
3 4 20\\
4 5 20\\
5 1 20\\

Output\\
0 25 60 40 20

Input\\
6 7 13 22\\
1 2 13\\
2 3 13\\
1 4 22\\
3 4 13\\
4 5 13\\
5 6 13\\
6 1 13\\

Output\\
0 13 26 39 26 13\\
\end{quote}

\subsection{Expert Summary}
\label{app:expert}

However, we can assume that an expert would already know what a minimum spanning tree is. Thus, we can remove this detailed description of an MST.

\begin{quote}
You are given a connected graph of $n$ nodes and $m$ bidirectional edges.
For each node $p = 1, 2, \dots, n$, you need to find a minimum spanning tree. Then output the minimum cost required to travel between node $1$ and node $p$.

-----Input-----

The first line of the input contains four integers $n$, $m$, $a$ and $b$ ($2 \leq n \leq 70$, $n - 1 \leq m \leq 200$, $1 \leq a < b \leq 10^7$) — the number of nodes and edges in the graph, and two possible travel times. Each of the following lines contains three integers $u, v, c$ ($1 \leq u, v \leq n$, $u \neq v$, $c \in \{a, b\}$) denoting an edge between the nodes $u$ and $v$, which has cost $c$.

You can assume that the graph is connected and has no loops or multiedges.

-----Output-----

Output a single line containing $n$ integers. The $p$-th of them should denote the minimum possible post required to travel from $1$ to $p$ after the selected edges are abandoned. Note that for each $p$ you can abandon a different set of edges.

-----Examples-----

Input\\
5 5 20 25\\
1 2 25\\
2 3 25\\
3 4 20\\
4 5 20\\
5 1 20\\

Output\\
0 25 60 40 20

Input\\
6 7 13 22\\
1 2 13\\
2 3 13\\
1 4 22\\
3 4 13\\
4 5 13\\
5 6 13\\
6 1 13\\

Output\\
0 13 26 39 26 13\\
\end{quote}

\section{Prompt templates}
\label{app:prompts}

\paragraph{Studio21}
Here is our template for Studio21. To see examples of summaries produced by Studio21AI's model along with the code generated for those summaries, refer to Appendix \ref{app:studio21-gen-code} and \ref{app:studio21-generated-code2}.

\begin{quote}
    The following sentences contain computer science jargon. Rewrite them using simple words.\\
    Jargon: <ORIGINAL>\\
    Simple: <SUMMARY>\\\\
    Jargon: <ORIGINAL>\\
    Simple: <SUMMARY>\\\\
    Jargon: <ORIGINAL>\\
    Simple: <SUMMARY>\\\\
    Jargon: <ORIGINAL>\\
    Simple:
\end{quote}

The few-shot examples were chosen randomly from the human generated expert summaries.

\paragraph{GPT3}
Here is our template for GPT3. To see examples of summaries produced by GPT3 and the code generated for those summaries refer to Appendix \ref{app:gpt-generated-code}.

\begin{quote}
    Summarize the following paragraph:
    Original: <ORIGINAL>\\
    Summary: <SUMMARY>\\\\
    Original: <ORIGINAL>\\
    Summary: <SUMMARY>\\\\
    Original: <ORIGINAL>\\
    Summary: <SUMMARY>\\\\
    Original: <ORIGINAL>
    Summary:
\end{quote}

\paragraph{Codex}
Here is our default template for Codex, which is used when there is no starter code provided. When there is starter code provided the docstring remains the same but the code after the doc string will be what is provided.

\begin{quote}
    Python3\\
    """\\
    <PROBLEM DESCRIPTION>\\
    """\\
    def code():
\end{quote}

\section{Strict Accuracy}
\label{app:accuracy}

Strict Accuracy (SAcc) is the percentage of problems that passed every test case. The formula to calculate SAcc is given below:

\begin{align}
    \textit{strict acc} &:= \frac{\textit{problems with 100\% accuracy}}{\textit{total number of problems}}
\end{align}

Given that, we are only generating one code solution for each problem our strict accuracy is comparable to \cite{chen2021codex}'s metric $raw\;pass @1$.

\section{Codex Configuration} 
\label{app:codex}

We did a small test with $75$ summaries to find our hyper-parameters for Codex. We set temperature to $0$, topP to $1$, frequency penalty to $0.2$, and presence penalty to $0$. We did not provide few-shot examples to Codex since we want to see if summarization only could improve the performance of the Codex model.

\section{Worst Problems and Statistics}
\label{app:worst_probs}
Using the test case labels as defined in section \ref{sec:exp-setup} we defined a test case as getting worse if it's label (result) was lower. Then we defined a problem as worse if \textit{every} test case had a lower label. Our methodology behind this was, if we removed problems that had a worse accuracy, then it would be a non-trivial result that accuracy improved. Also, if we removed problems with worse accuracy, then a problem that originally had all $0$ labels (all False test cases) would score the same if the summary had all $-1$ labels (runtime error) or a $-2$ (syntax error). So, we removed problems which every test case performed worse, to see if removing these outliers would improve results. You can see the overall breakdown of each split in table \ref{tab:number-of-probs}.

\begin{table}[t!]
\centering
\small
\resizebox{0.95\linewidth}{!}{

\begin{tabular}{@{}lll|l|l@{}}
\toprule
Summary & Difficulty & AP & EWPR & BWPR\\
 \toprule
 \multirow{3}{*}{Basic}
 & Introductory & 145 & 141 & 144 \\
 & Interview & 123 & 113 & 123 \\
 & Competition & 105 & 100 & 105 \\
 \midrule
 \multirow{3}{*}{Expert}
 & Introductory & 145 & 140 & 144 \\
 & Interview & 123 & 116 & 123 \\
 & Competition & 105 & 100 & 105 \\
 \midrule
  \multirow{3}{*}{StudioAI21}
 & Introductory & 215 & 187 & - \\
 & Interview & 627 & 558 & - \\
 & Competition & 659 & 578 & - \\
 \midrule
  \multirow{3}{*}{GPT3}
 & Introductory & 194 & 180 & - \\
 & Interview & 266 & 242 & - \\
 & Competition & 244 & 220 & - \\
 \bottomrule
\end{tabular}
}
\caption{These are the numbers of problems in each split of the dataset. For GPT and Studio21 we did not look at problems that were worse or same for both experiments because there was insignificant overlap between the two experiments.}
\label{tab:number-of-probs}
\end{table}

\section{Average length of Problems and Solutions}
\label{app:avg_length}

Table \ref{tab:aggregate-results} represents the statistics for average length of problems and solutions for original and summarized prompts.

\begin{table*}[t!]
\centering
\small
\resizebox{0.95\linewidth}{!}{

\begin{tabular}{@{}l|l|l|l|l|l@{}}
\toprule
Experiment & Original Len & Summary Len & Orig Code Len & Summary Code Len & Code Solution Len\\
 \toprule
Summary & 1147 & 937 & 339 & 349 & 671 \\
 \midrule
Expert &  1147 & 869 & 339 & 343 & 671 \\
  \midrule
GPT & 1386 & 1011 & 437 & 392 & 748 \\
  \midrule
StudioAI21 & 1646 & 1114 & 602 & 473 & 721 \\
 \bottomrule
\end{tabular}
}
\caption{The average length of the original/summarized prompt and generated code. The average length of the code solutions is the average len of the solutions provided by the creators of the APPS dataset. A problem could have one or multiple solutions. The length is reported in characters.}
\label{tab:aggregate-results}
\end{table*}

\section{Abbreviated Synthetic Results}
\label{app:abb_syn_results}

In table \ref{tab:condensed-synthetic-results}, we show the results for our synthetic summaries when taking the top 500 and 1000 summaries for GPT3 and StudioAI21, respectively. In our initial experiment, this was the amount of problems we tested for each model. However, in our final experiment we changed our configurations and generated more problems. For a comparison, we took the top performing summaries and and reported those results.

\begin{table}[t!]
\centering
\small
\resizebox{\linewidth}{!}{

\begin{tabular}{@{}llcc|cc@{}}
\toprule
\multirow{2}{*}{Model} & \multirow{2}{*}{Difficulty} & \multicolumn{2}{c}{AP} & \multicolumn{2}{c}{EWPR} \\
\cmidrule(lr){3-4}\cmidrule(lr){5-6}
~& ~ & Baseline & Proposed & Baseline & Proposed \\ 
 \toprule
 \multirow{2}{*}{GPT-3}
 & Introductory & \textbf{41.97} & 38.86 & 41.11 & \textbf{41.67} \\
 & Interview & 25.27 & \textbf{27.47} & 24.86 & \textbf{28.25} \\
 & Competition & 4.80 & \textbf{6.40} & 4.88 & \textbf{6.50} \\
 \midrule
 & Weighted Average & \textbf{26.60} & \textbf{26.60} & 25.83 & \textbf{27.71} \\
 \midrule
 \multirow{3}{*}{Studio21}
 & Introductory & \textbf{39.91} & 31.92 & \textbf{39.25} & 36.56 \\
 & Interview & \textbf{15.97} & 14.50 & 13.23 & \textbf{15.47} \\
 & Competition & \textbf{2.57} & \textbf{2.57} & \textbf{2.71} & \textbf{2.71} \\
 \midrule
 & Weighted Average & \textbf{16.90} & 14.50 & 15.10 & \textbf{15.64} \\
 \bottomrule
\end{tabular}
}
\caption{Results when taking the top 500 GPT problems and top 1000 Studio problems}
\label{tab:condensed-synthetic-results}
\end{table}

\section{Generated Code}
\label{app:generated-code}

\begin{figure*}[ht]
    \centering
    \includegraphics[width=\linewidth]{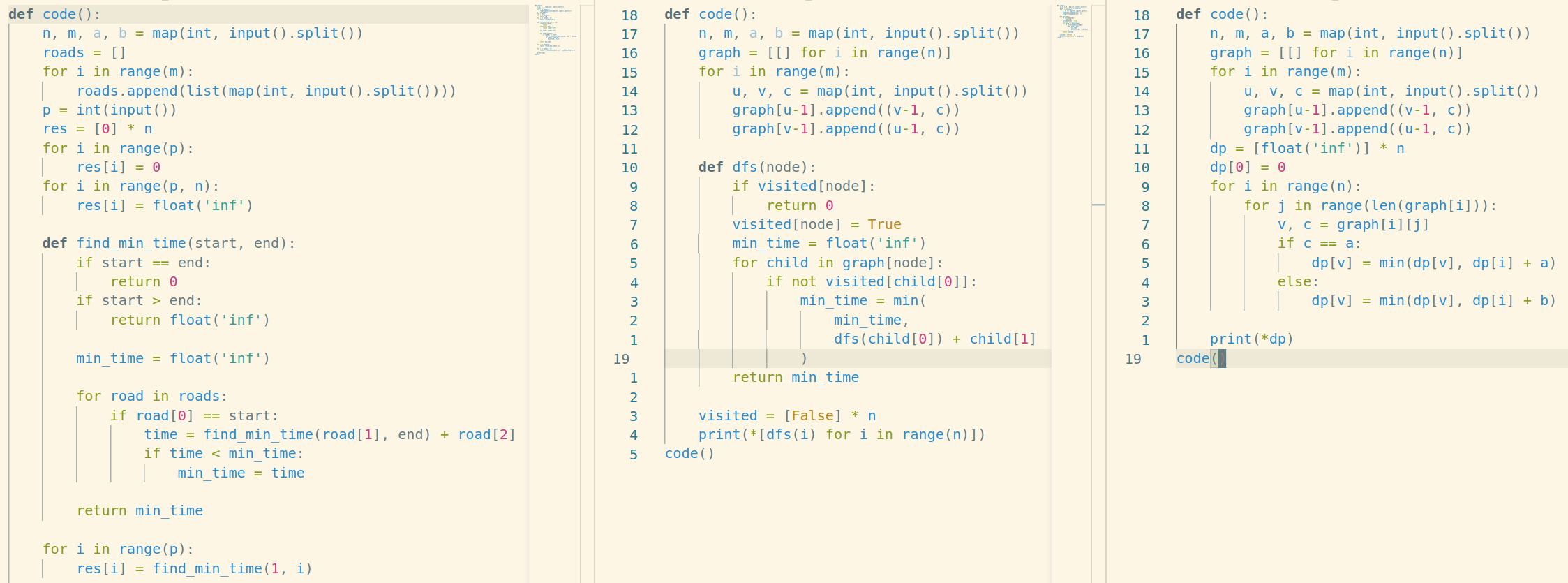}
    \caption{On the far left is the code generated by the original prompt. The middle is the code generated by the expert summary. The right is the code generated by the basic summary.}
    \label{fig:comp-2000-code}
\end{figure*}

In figure \ref{fig:comp-2000-code} is the code that was generated for the example mentioned in Appendix \ref{app:example_fake} and \ref{app:expert}. Given that the Codex model was prompted with the $def\:code():$ the model did not generate that function definition or the call to that function. That was added in afterwards, but everything inside that function was generated by Codex. The originally generated code (far left) fails with a $-1$ because it did not take in the input correctly. It added in another line \textit{p = int(input())}, which most likely refers to the $p$ mentioned in the original text. The expert summary generated code (middle) fails every test case. The basic summary generated code (right) passed $16/19\,(84\%)$ test cases and was the only code to pass at least $1$ test case.

\section{StudioAI21 Generated Code}
\label{app:studio21-gen-code}
Below is an example of a competition problem where StudioAI21 summarized the prompt too much but Codex was still able to produce viable code. Here is the original prompt:

\begin{quote}
Cengiz recently learned Fibonacci numbers and now he is studying different algorithms to find them. After getting bored of reading them, he came with his own new type of numbers that he named XORinacci numbers. He defined them as follows:   $f(0) = a$;  $f(1) = b$;  $f(n) = f(n-1) \oplus f(n-2)$ when $n > 1$, where $\oplus$ denotes the bitwise XOR operation. 

You are given three integers $a$, $b$, and $n$, calculate $f(n)$.

You have to answer for $T$ independent test cases.

-----Input-----

The input contains one or more independent test cases.

The first line of input contains a single integer $T$ ($1 \le T \le 10^3$), the number of test cases.

Each of the $T$ following lines contains three space-separated integers $a$, $b$, and $n$ ($0 \le a, b, n \le 10^9$) respectively.

-----Output-----

For each test case, output $f(n)$.

-----Example-----\\
Input\\
3\\
3 4 2\\
4 5 0\\
325 265 1231232\\

Output\\
7\\
4\\
76\\

-----Note-----

In the first example, $f(2) = f(0) \oplus f(1) = 3 \oplus 4 = 7$.
\end{quote}

Here is the summary that StudioAI21 generated:

\begin{quote}
 You are given three integers $a$, $b$, and $n$. Calculate $f(n)$. 
 
-----Input-----

The input contains one or more independent test cases.

The first line of input contains a single integer $T$ ($1 \le T \le 10^3$), the number of test cases.

Each of the $T$ following lines contains three space-separated integers $a$, $b$, and $n$ ($0 \le a, b, n \le 10^9$) respectively.

-----Output-----

For each test case, output $f(n)$.

-----Example-----\\
Input\\
3\\
3 4 2\\
4 5 0\\
325 265 1231232\\

Output\\
7\\
4\\
76\\

-----Note-----

In the first example, $f(2) = f(0) \oplus f(1) = 3 \oplus 4 = 7$.
\end{quote}

\begin{figure*}[ht]
    \centering
    \includegraphics[width=\linewidth]{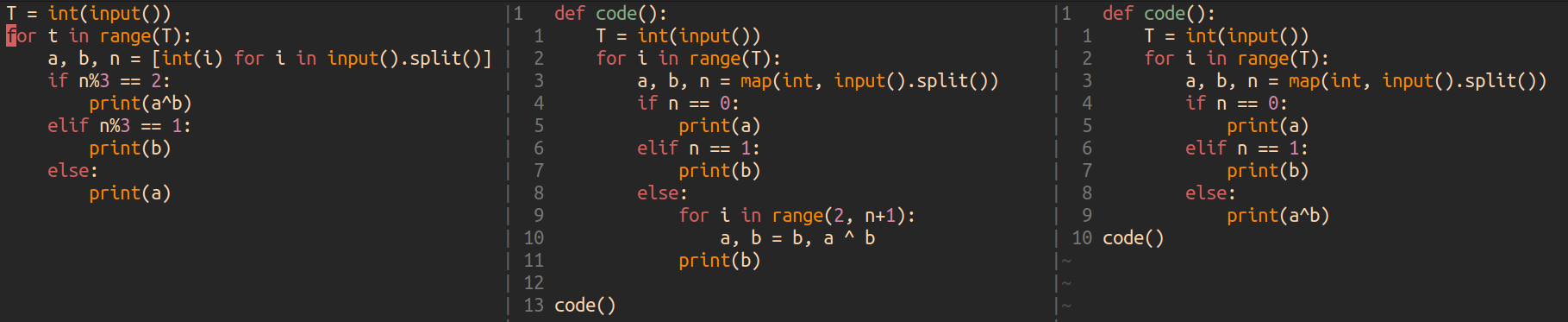}
    \caption{The left is the code generated using the original prompt. The right is the code generated when using the StudioAI21 generated summary.}
    \label{fig:comp-2213}
\end{figure*}

Because any input/output examples provided by the prompt are appended to the summary, Codex was able to figure out the pattern in the problem and generate code that was almost correct. In figure \ref{fig:comp-2213}, the solution (left) used the pattern in the problem and simplify by taking $n \mod 3$. The Studio21 summary code (right) recognizes this pattern but erroneously does not take the modulus of the number. The original code (center) also makes the same mistake by not taking the modulus, but also brute forces the answer. This shows that the model did not recognize the pattern in this problem because of the superfluous details. Even though Studio21 might have summarized too much, the model was still able to make an improvement and understand the pattern in the problem more. 

\section{StudioAI21 Generated Code}
\label{app:studio21-generated-code2}
Here is an example of a summary made by StudioAI21 where the qualitative aspect of the code but it still failed. Here is the original prompt:

\begin{quote}
Given is a tree G with N vertices.
The vertices are numbered 1 through N, and the i-th edge connects Vertex $a_i$ and Vertex $b_i$.
Consider painting the edges in G with some number of colors.
We want to paint them so that, for each vertex, the colors of the edges incident to that vertex are all different.
Among the colorings satisfying the condition above, construct one that uses the minimum number of colors.

-----Constraints-----\\
 -  $2 \le N \le 10^5$\\
 -  $1 \le a_i < b_i \le N$\\
 - All values in input are integers.\\
 - The given graph is a tree.\\

-----Input-----\\
Input is given from Standard Input in the following format:\\
N\\
$a_1$ $b_1$\\
$a_2$ $b_2$\\
$\vdots$\\
$a_{N-1}$ $b_{N-1}$\\

-----Output-----\\
Print N lines.\\
The first line should contain K, the number of colors used.\\
The (i+1)-th line $(1 \le i \le N-1)$ should contain $c_i$, the integer representing the color of the i-th edge, where $1 \le c_i \le K$ must hold.\\
If there are multiple colorings with the minimum number of colors that satisfy the condition, printing any of them will be accepted.

-----Sample Input-----\\
3\\
1 2\\
2 3\\

-----Sample Output-----\\
2\\
1\\
2\\
\end{quote}

Here is the summary that StudioAI21 generated:

\begin{quote}
     You are given a tree with N vertices numbered from 1 to N. You need to find the minimum possible number of colors to paint the edges such that vertices are painted with different color.
     -----Constraints-----\\
 -  $2 \le N \le 10^5$\\
 -  $1 \le a_i < b_i \le N$\\
 - All values in input are integers.\\
 - The given graph is a tree.\\

-----Input-----\\
Input is given from Standard Input in the following format:\\
N\\
$a_1$ $b_1$\\
$a_2$ $b_2$\\
$\vdots$\\
$a_{N-1}$ $b_{N-1}$\\

-----Output-----\\
Print N lines.\\
The first line should contain K, the number of colors used.\\
The (i+1)-th line $(1 \le i \le N-1)$ should contain $c_i$, the integer representing the color of the i-th edge, where $1 \le c_i \le K$ must hold.\\
If there are multiple colorings with the minimum number of colors that satisfy the condition, printing any of them will be accepted.

-----Sample Input-----\\
3\\
1 2\\
2 3\\

-----Sample Output-----\\
2\\
1\\
2\\
\end{quote}

\begin{figure*}[ht]
    \centering
    \includegraphics[width=\linewidth]{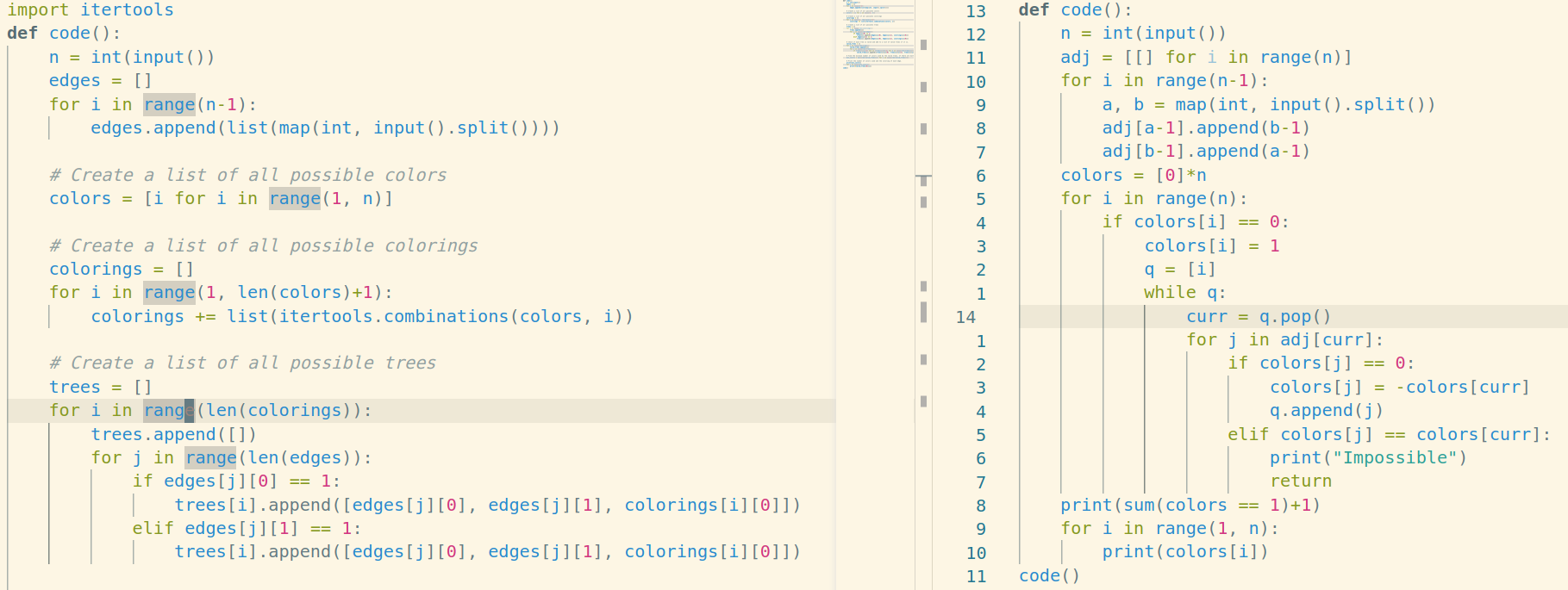}
    \caption{The left is the code generated using the original prompt. The right is the code generated when using the StudioAI21 generated summary.}
    \label{fig:interview-1273}
\end{figure*}

In \ref{fig:interview-1273} the left is the original solution which fails with a $-2$ because the runtime of the algorithm is exponential. Note that it tries to create a list of all possible edge colorings which is $\mathcal{O}(2^N)$. The right is the code produced when using the StudioAI21 summary. You can see that this code is much closer to solving the problem and produces an efficient algorithm. However, this fails with a $-2$ because it tries to print the sum of a boolean (near the end before the last for loop). Which fails in python because a bool is not iterable.

Here is a problem where StudioAI21's summary increased the accuracy to $100\%$. Here is the original prompt:
\begin{quote}
Polycarpus has a sequence, consisting of n non-negative integers: $a_1, a_2, ..., a_{n}$.

Let's define function f(l, r) (l, r are integer, $1 \le l \le r \le n$) for sequence a as an operation of bitwise OR of all the sequence elements with indexes from l to r. Formally: f(l, r) = $a_{l}|a_{l} + 1|...|a_{r}.$ 

Polycarpus took a piece of paper and wrote out the values of function f(l, r) for all l, r (l, r are integer, $1 \le l \le r \le n$). Now he wants to know, how many distinct values he's got in the end. 

Help Polycarpus, count the number of distinct values of function f(l, r) for the given sequence a.

Expression x|y means applying the operation of bitwise OR to numbers x and y. This operation exists in all modern programming languages, for example, in language C++ and Java it is marked as "|", in Pascal — as "or".

-----Input-----

The first line contains integer n $(1 \le n \le 10^5)$ — the number of elements of sequence a. The second line contains n space-separated integers $a_1, a_2, ..., a_{n}$ $(0 \le a_{i} \le 10^6)$ — the elements of sequence a.

-----Output-----

Print a single integer — the number of distinct values of function f(l, r) for the given sequence a.

Please, do not use the lld specifier to read or write 64-bit integers in C++. It is preferred to use cin, cout streams or the I64d specifier.

-----Examples-----\\
Input\\
3\\
1 2 0\\

Output\\
4\\
Input\\
10\\
1 2 3 4 5 6 1 2 9 10\\

Output\\
11

-----Note-----

In the first test case Polycarpus will have 6 numbers written on the paper: f(1, 1) = 1, f(1, 2) = 3, f(1, 3) = 3, f(2, 2) = 2, f(2, 3) = 2, f(3, 3) = 0. There are exactly 4 distinct numbers among them: 0, 1, 2, 3.
\end{quote}

Here is the summary that StudioAI21 generated. Not exactly as we would expect as the prompt still mentions the fictional \textit{Polycarpus}.

\begin{quote}
 Polycarpus has a sequence, consisting of n non-negative integers: $a_1, a_2, ..., a_{n}$. Let's define function f(l, r) (l, r are integer, 1   l   r   n) for sequence a as an operation of bitwise OR of all the sequence elements with indexes from l to r. Formally: f(l, r) = $a_{l} | a_{l} + 1 | ...$

-----Input-----

The first line contains integer n $(1 \le n \le 10^5)$ — the number of elements of sequence a. The second line contains n space-separated integers $a_1, a_2, ..., a_{n}$ $(0 \le a_{i} \le 10^6)$ — the elements of sequence a.

-----Output-----

Print a single integer — the number of distinct values of function f(l, r) for the given sequence a.

Please, do not use the lld specifier to read or write 64-bit integers in C++. It is preferred to use cin, cout streams or the I64d specifier.

-----Examples-----\\
Input\\
3\\
1 2 0\\

Output\\
4\\
Input\\
10\\
1 2 3 4 5 6 1 2 9 10\\

Output\\
11

-----Note-----

In the first test case Polycarpus will have 6 numbers written on the paper: f(1, 1) = 1, f(1, 2) = 3, f(1, 3) = 3, f(2, 2) = 2, f(2, 3) = 2, f(3, 3) = 0. There are exactly 4 distinct numbers among them: 0, 1, 2, 3.
\end{quote}

\begin{figure*}[ht]
    \centering
    \includegraphics[width=\linewidth]{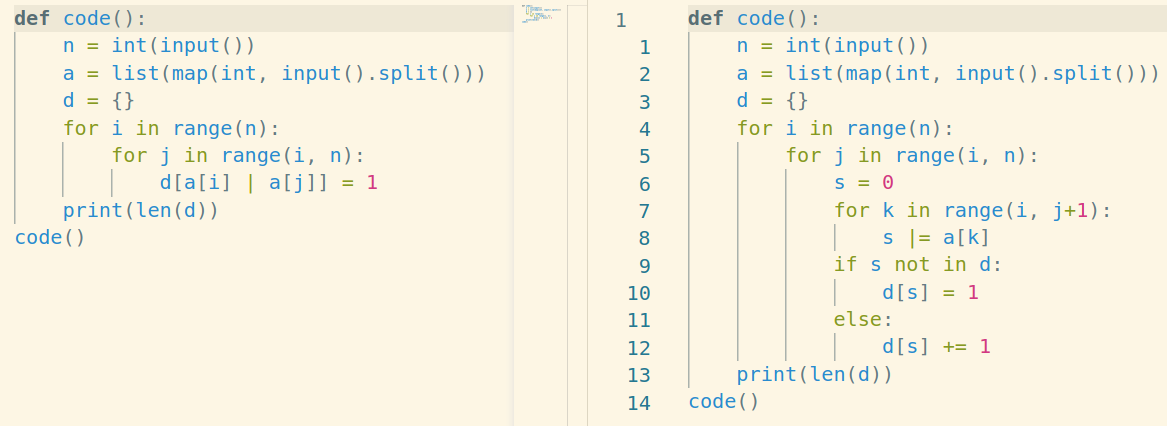}
    \caption{The left is the code generated using the original prompt. The right is the code generated when using the StudioAI21 generated summary.}
    \label{fig:comp-2174}
\end{figure*}

In \ref{fig:comp-2174} the left is the original solution which gets $77\%$ accuracy. The right is the summary code which gets $100\%$ accuracy.

\section{GPT Generated Summaries}
\label{app:gpt-generated-code}
Here are two summaries where GPT perfectly summarized the prompt and gave a concise description of what the task was. In both cases the original prompt did not have 100\% accuracy but the summarized prompt did have 100\% accuracy.

Here is the original prompt for the first question:
\begin{quote}
    Božo is a strange little boy. Every day he tires his friends with strange questions. Today’s question is: how many integers in the interval $[A, B]$ are there such that the sum of their digits is $S$, and which is the smallest such number?

Write a program that answers Božo’s question so that he can get some sleep.

-----Input-----\\
The input contains three integers $A$, $B$ and $S$ ($1 \le A \le B < 10^{15}$, $1 \le S \le 135$).

-----Output-----\\
The first line should contain the number of integers in the interval with the digit sum equal to $S$.

The second line should contain the smallest such integer.

The input data will guarantee that the first number is at least $1$.

-----Examples-----\\
Sample Input 1:\\
1 9 5\\
Sample Output 1:\\
1\\
5\\

Sample Input 2:\\
1 100 10\\
Sample Output 2:\\
9\\
19\\
\end{quote}

Here is the summary that GPT Generated:
\begin{quote}
Write a program that calculates the smallest integer in the given interval whose sum of digits is equal to the given sum.

-----Input-----\\
The input contains three integers $A$, $B$ and $S$ ($1 \le A \le B < 10^{15}$, $1 \le S \le 135$).

-----Output-----\\
The first line should contain the number of integers in the interval with the digit sum equal to $S$.

The second line should contain the smallest such integer.

The input data will guarantee that the first number is at least $1$.

-----Examples-----\\
Sample Input 1:\\
1 9 5\\
Sample Output 1:\\
1\\
5\\

Sample Input 2:\\
1 100 10\\
Sample Output 2:\\
9\\
19\\
\end{quote}

\begin{figure*}[ht]
    \centering
    \includegraphics[width=\linewidth]{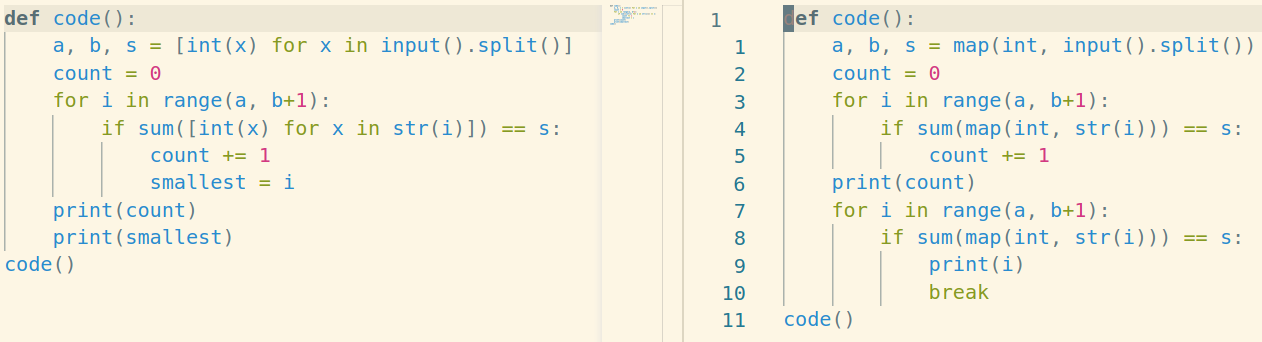}
    \caption{The left is the code generated using the original prompt. The right is the code generated when using the GPT3 generated summary.}
    \label{fig:comp-3142}
\end{figure*}

In \ref{fig:comp-3142} you can see the original code on the left and the summary code on the right. There is a subtle difference but it's that difference that improved the problem from $33\%$ accuracy to $100\%$.

Here is the original prompt for another question.
\begin{quote}
    Professor GukiZ makes a new robot. The robot are in the point with coordinates $(x_1, y_1)$ and should go to the point $(x_2, y_2)$. In a single step the robot can change any of its coordinates (maybe both of them) by one (decrease or increase). So the robot can move in one of the 8 directions. Find the minimal number of steps the robot should make to get the finish position.

-----Input-----

The first line contains two integers $x_1, y_1$ $( - 10^9 \le x_1, y_1 \le 10^9)$ — the start position of the robot.

The second line contains two integers $x_2, y_2$ $( - 10^9 \le x_2, y_2 \le 10^9)$ — the finish position of the robot.

-----Output-----

Print the only integer d — the minimal number of steps to get the finish position.

-----Examples-----\\
Input\\
0 0\\
4 5

Output\\
5

Input\\
3 4\\
6 1

Output\\
3

-----Note-----

In the first example robot should increase both of its coordinates by one four times, so it will be in position (4, 4). After that robot should simply increase its y coordinate and get the finish position.

In the second example robot should simultaneously increase x coordinate and decrease y coordinate by one three times.
\end{quote}

Here is the summary that GPT3 generated:

\begin{quote}
The robot can move in one of the 8 directions. Find the minimal number of steps the robot should make to get the finish position.

-----Input-----

The first line contains two integers $x_1, y_1$ $( - 10^9 \le x_1, y_1 \le 10^9)$ — the start position of the robot.

The second line contains two integers $x_2, y_2$ $( - 10^9 \le x_2, y_2 \le 10^9)$ — the finish position of the robot.

-----Output-----

Print the only integer d — the minimal number of steps to get the finish position.

-----Examples-----\\
Input\\
0 0\\
4 5

Output\\
5

Input\\
3 4\\
6 1

Output\\
3

-----Note-----

In the first example robot should increase both of its coordinates by one four times, so it will be in position (4, 4). After that robot should simply increase its y coordinate and get the finish position.

In the second example robot should simultaneously increase x coordinate and decrease y coordinate by one three times.
\end{quote}

\begin{figure*}[ht]
    \centering
    \includegraphics[width=\linewidth]{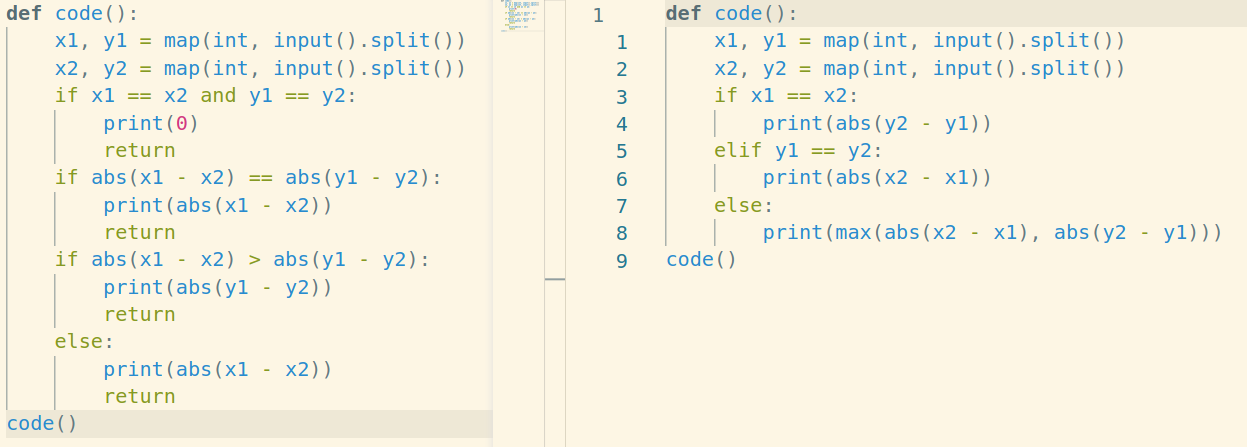}
    \caption{The left is the code generated using the original prompt. The right is the code generated when using the GPT3 generated summary.}
    \label{fig:inter-0202}
\end{figure*}

In \ref{fig:inter-0202} you can see the original code on the left and the summary code on the right. There is a subtle difference but it's that difference that improved the problem from $20\%$ accuracy to $100\%$.

\section{Human Generated Instructions}
\label{app:instructions}
The section below was given to each crowd worker as instructions to follow when creating the regular and expert summaries.

\subsection{Summarization}

Create a file called $summary.txt$ this will contain your summary of the prompt. It's recommended that you copy the $question.txt$ file into the $summary.txt$ file then starting from the top of the prompt follow the steps and remove words/lines as necessary.

These are the rough steps for making a summary. Following these steps will create the most consistency in our dataset. However, you should summarize as you see fit. First, read through the prompt and understand what it's asking, then follow these steps to help create a summary.

\begin{enumerate}
    \item Directly state what is given in the problem.
    \begin{itemize}
        \item Most problems start by setting the scene, to help humans understand.
        \item Start the problems by explicitly telling the model what the input is. 
        \item \textit{You are given \ldots}
    \end{itemize}
    \item Remove any $notes$ given in the prompt.
    \begin{itemize}
        \item They are usually reemphasizing points, which is redundant and not needed in the summary.
        \item This includes the $-Notes-$ section at the bottom of the file.
        \item If there is pertinent information given from a note, include it in the prompt without describing it as a note.
    \end{itemize}
    \item Remove any text in parenthesis.
    \begin{itemize}
        \item Most of the text in parenthesis is repeating the information that precede them.
        \item If the text in parenthesis provides more context or information, then remove the preceding text.
        \item Keep any parenthesis if it is describing constraints, such as the minimum and maximum values for the input etc...
    \end{itemize}
    \item Remove any made up people, places, things, etc...
    \begin{itemize}
        \item These abstractions are made to help humans understand but confuse the model.
        \item The prompts often mention things like $Codefortia$ or $Polycarp$, try to replace these with the word $you$.
        \item Any text visualizing what the problem is asking, should be removed.
    \end{itemize}
    \item If the $Input$ or $Output$ section reference an abstraction they should be changed.
    \begin{itemize}
        \item Overall, these sections are fine. However, if they mentioned something you removed in the previous steps, they should be changed to reflect that.
        \item If these sections repeat themselves remove any redundancies.
        \item In most cases these sections will be left alone.
    \end{itemize}
\end{enumerate}

\subsection{Expert Summary}
Create a file called $expert.txt$ this will contain an expert summary of the prompt. It's recommended that you copy the $summary.txt$ file into the $expert.txt$ file then starting from the top of the prompt remove words/lines as necessary. You should aim for the expert prompt to be $2-4$ lines.

Imagine you are describing the prompt to a senior software engineer. What else could you trim out? The difference between the original and expert summary, is the original summary may include something obvious, whereas the expert solution should be the absolute bare minimum. To create $summary.txt$ you want to remove superfluous details from the original prompt. To create $expert.txt$ you want to remove details that an expert would find obvious, from the summary.

For example, in problem 2000 (which is competitive difficulty) the summary mentions '\textit{It will be possible to travel between each pair of nodes \ldots, and the sum of times \ldots will be the minimum possible}'. This process is describing a minimum spanning tree so you can just say '\textit{Find a minimum spanning tree}'.

Also, if the prompt included an example and subsequent explanation, that should remain in the summary but should be removed from the expert summary. An expert already understands the problem and does not need any extra explanation. You should still keep the $-Examples-$ section.



\paragraph{Takeaways}
\begin{itemize}
    \item Removing made up people, places, and things from the prompt improved the quality of code generated.
    \item The optimal summarization depends on the difficulty of the problem. 
    \item Synthetically generate summaries were close to maintaining accuracy.
    \item With more rigorous instructions, human summaries could be made with less noise which would further improve synthetic summary generation.
\end{itemize}

\section{Superfluous Information Confusing the Model}
\label{app:extra-information}
Here is an example of an interview level string problem where the original prompt got $0\%$ and both human generated summaries got $100\%$ accuracy. The question wants you to write code that will return the number of unique character in the given string.

\subsection{Original Prompt}
\begin{quote}
    You have initially a string of N characters, denoted by A1,A2...AN. You have to print the size of the largest subsequence of string A such that all the characters in that subsequence are distinct ie. no two characters in that subsequence should be same.

A subsequence of string A is a sequence that can be derived from A by deleting some elements  and without changing the order of the remaining elements.

-----Input-----
First line contains T, number of testcases. Each testcase consists of a single string in one line. Each character of the string will be a small alphabet(ie. 'a' to 'z').

-----Output-----
For each testcase, print the required answer in one line.

-----Constraints-----\\
- $1 \leq T \leq 10$\\
- Subtask 1 (20 points):$1 \leq N \leq 10$\\
- Subtask 2 (80 points):$1 \leq N \leq 105$\\

-----Example-----\\
Input:\\
2\\
abc\\
aba\\

Output:
3\\
2

-----Explanation-----
For first testcase, the whole string is a subsequence which has all distinct characters.

In second testcase, the we can delete last or first 'a' to get the required subsequence.
\end{quote}

\subsection{Basic Summary}

\begin{quote}
    You are given N string. You have to identify the duplicates and print the length of the new string as a combination of unique characters only.

-----Input-----
First line contains T, number of testcases. Each testcase consists of a single string in one line. Each character of the string will be a small alphabet(ie. 'a' to 'z').

-----Output-----
For each testcase, print the required answer in one line.

-----Constraints-----\\
- $1 \leq T \leq 10$\\
- Subtask 1 (20 points):$1 \leq N \leq 10$\\
- Subtask 2 (80 points):$1 \leq N \leq 105$\\

-----Example-----\\
Input:\\
2\\
abc\\
aba\\

Output:
3\\
2

\end{quote}

\subsection{Expert Summary}
\begin{quote}
    You have to remove duplicates and print the length of unique characters of the given string.

-----Input-----
First line contains T, number of testcases. Each testcase consists of a single string in one line. Each character of the string will be a small alphabet(ie. 'a' to 'z').

-----Output-----
For each testcase, print the required answer in one line.

-----Constraints-----\\
- $1 \leq T \leq 10$\\
- Subtask 1 (20 points):$1 \leq N \leq 10$\\
- Subtask 2 (80 points):$1 \leq N \leq 105$\\

-----Example-----\\
Input:\\
2\\
abc\\
aba\\

Output:
3\\
2

\end{quote}

\subsection{Generated Code}
The original code (left) does not accomplish the task but rather prints the count of the most frequent character. The model was unable to distinguish what the task was given the verbose prompt. However, the basic and expert summaries make the task clear and the model produces the same code. Which properly solves the challenge.

\begin{figure*}[ht]
    \centering
    \includegraphics[width=\linewidth]{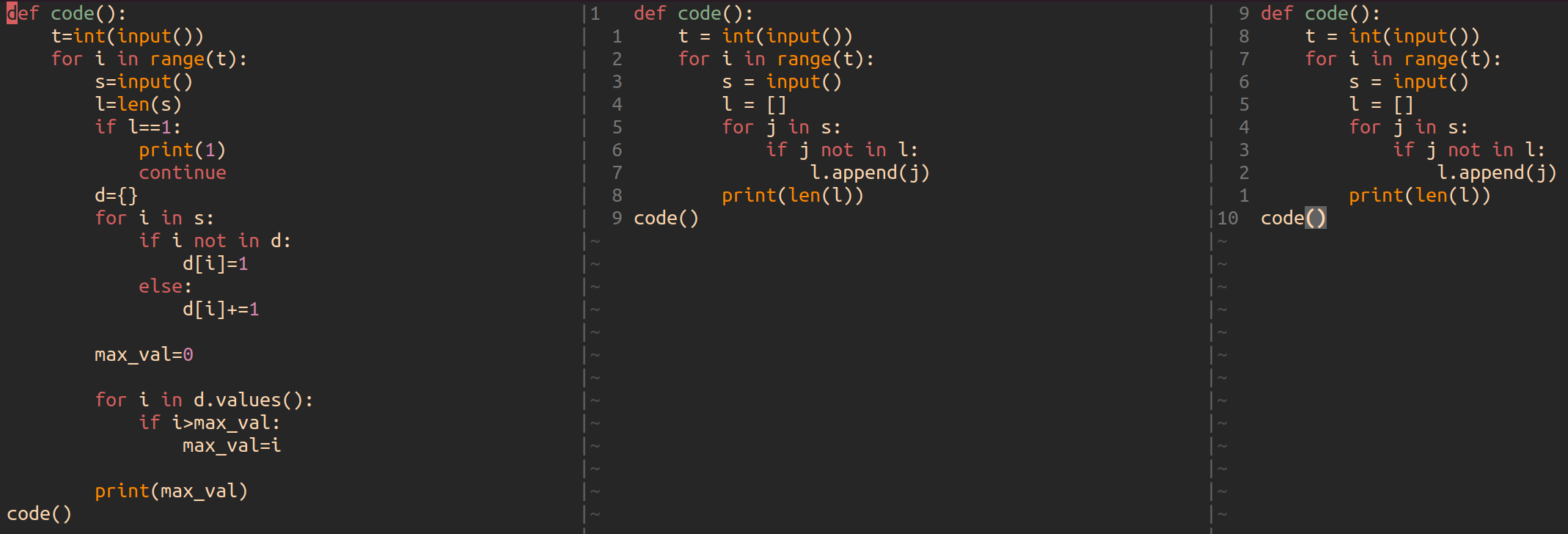}
    \caption{The left is the code generated by the original prompt. The middle is the code generated by the expert summary. The right is the code generated by the basic summary.}
    \label{fig:interview-1398}
\end{figure*}

\section{Made Up Information Confusing the Model}
\label{app:fake-information}
Here is an example of an interview level problem where the original prompt got $0\%$ and the expert generated summary got $100\%$ accuracy. 

\subsection{Original Prompt}

\begin{quote}
    The chef was searching for his pen in the garage but he found his old machine with a display and some numbers on it. If some numbers entered then some different output occurs on the display. Chef wants to crack the algorithm that the machine is following.
Example to identify the pattern :\\

Input\,\,\,\,                          Output\\
9     \,     \,\,\,\,\,\,\,\,\,                       36\\
5      \,   \,\,\,\,\,\,\,\,\,                          10\\
1       \,   \,\,\,\,\,\,\,\,\,                          0\\
2        \,     \,\,\,\,\,\,\,\,\,                       1\\

-----Input:-----\\
- First-line will contain $T$, the number of test cases. Then the test cases follow. 
- Each test case contains a single line of input, $N$. 

-----Output:-----\\
For each test case, output in a single line answer as displayed on the screen.

-----Constraints-----\\
- $1 \leq T \leq 10^6$\\
- $1 \leq N \leq 10^6$

-----Sample Input:-----\\
1\\
7\\

-----Sample Output:-----\\
21
\end{quote}

\subsection{Expert Summary}

\begin{quote}
    
Write a code to print the average of the multiplication of a given number N with N-1 integer.
                                   1

-----Input:-----\\
- First-line will contain $T$, the number of test cases. Then the test cases follow. 
- Each test case contains a single line of input, $N$. 

-----Output:-----\\
For each test case, output in a single line answer as displayed on the screen.

-----Constraints-----\\
- $1 \leq T \leq 10^6$\\
- $1 \leq N \leq 10^6$

-----Sample Input:-----\\
1\\
7\\

-----Sample Output:-----\\
21
\end{quote}

\begin{figure*}[ht]
    \centering
    \includegraphics[width=\linewidth]{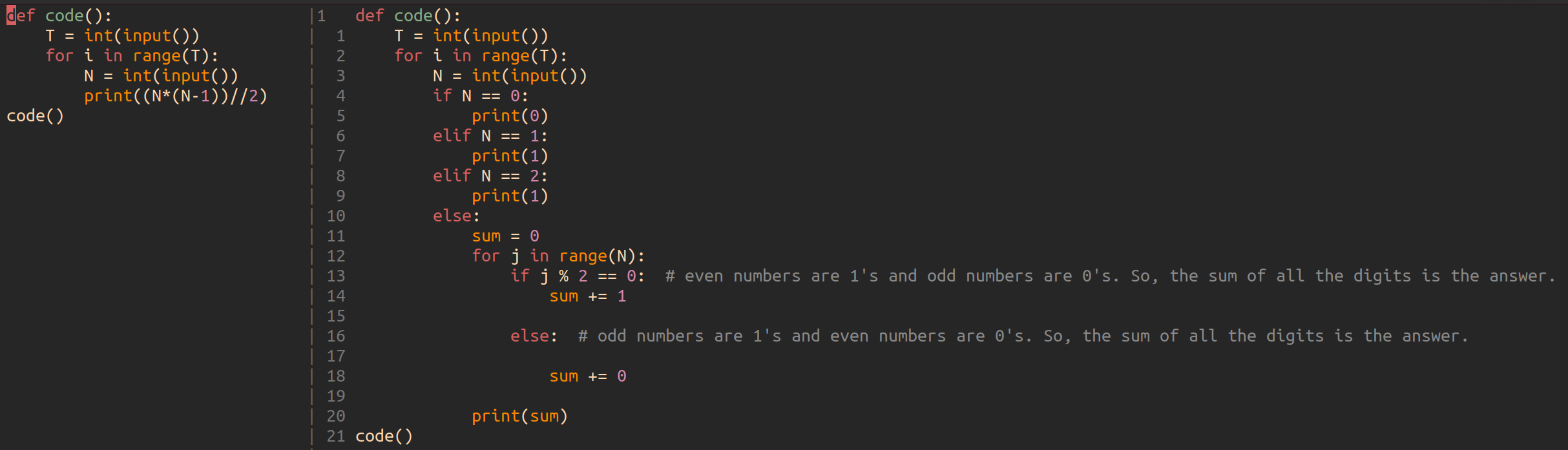}
    \caption{ The left is the code generated by the expert summary. The right is the code generated by the original prompt.}
    \label{fig:interview-1380}
\end{figure*}
\end{document}